\documentclass[conference]{IEEEtran}

\usepackage{cite}
\usepackage{amsmath,amssymb,amsfonts}
\usepackage{amssymb}
\usepackage{algorithmic}
\usepackage{graphicx}
\usepackage{textcomp}
\usepackage{xcolor}
\usepackage{subfigure}
\usepackage{amsthm}
\usepackage{mathrsfs}

\usepackage{pdfpages}
\usepackage{colortbl}
\definecolor{mygray}{gray}{0.85}
\usepackage{array}
\usepackage{caption}
\usepackage{setspace}
\usepackage{booktabs}
\usepackage[linesnumbered,boxed,ruled,commentsnumbered]{algorithm2e}

\usepackage[ruled,linesnumbered]{algorithm2e}

\usepackage{amsmath,bm}
\usepackage{cite}
\usepackage{amsmath,amssymb,amsfonts}
\usepackage{algorithmic}
\usepackage{amsthm}
\usepackage{graphicx}
\usepackage{textcomp}
\usepackage{xcolor}

\makeatletter

\newcommand{\Rmnum}[1]{\expandafter\@slowromancap\romannumeral #1@}
\makeatother

\begin{document}
\begin{spacing}{1.0}
%
\title{
\Huge
Cooperative Target Detection with AUVs:\\ A Dual-Timescale Hierarchical MARDL Approach 
}

\author{
\IEEEauthorblockN{Xueyao Zhang\IEEEauthorrefmark{1}, Bo Yang\IEEEauthorrefmark{1}, Zhiwen Yu\IEEEauthorrefmark{1}\IEEEauthorrefmark{2}, Xuelin Cao\IEEEauthorrefmark{3}, George C. Alexandropoulos\IEEEauthorrefmark{4}\IEEEauthorrefmark{5}, \\ M\'erouane Debbah\IEEEauthorrefmark{6}, and Chau Yuen\IEEEauthorrefmark{7}} 
\IEEEauthorblockA{\IEEEauthorrefmark{1}School of Computer Science, Northwestern Polytechnical University, Xi'an, Shaanxi, 710129, China} 
\IEEEauthorblockA{\IEEEauthorrefmark{2}Harbin Engineering University, Harbin, Heilongjiang, 150001, China} 
\IEEEauthorblockA{\IEEEauthorrefmark{3}School of Cyber Engineering, Xidian University, Xi'an, Shaanxi, 710071, China}
\IEEEauthorblockA{\IEEEauthorrefmark{4}Department of Informatics and Telecommunications, National and Kapodistrian University of Athens, 15784 Athens, Greece}
\IEEEauthorblockA{\IEEEauthorrefmark{5}Department of Electrical and Computer Engineering, University of Illinois Chicago, IL 60601, USA}
\IEEEauthorblockA{\IEEEauthorrefmark{6}Center for 6G Technology, Khalifa University of Science and Technology, P O Box 127788, Abu Dhabi, UAE}
\IEEEauthorblockA{\IEEEauthorrefmark{7}School of Electrical and Electronics Engineering, Nanyang Technological University, Singapore}
}


\maketitle

\begin{abstract}
Autonomous Underwater Vehicles (AUVs) have shown great potential for cooperative detection and reconnaissance. However, collaborative AUV communications introduce risks of exposure. In adversarial environments, achieving efficient collaboration while ensuring covert operations becomes a key challenge for underwater cooperative missions. In this paper, we propose a novel dual time-scale Hierarchical Multi-Agent Proximal Policy Optimization (H-MAPPO) framework. The high-level component determines the individuals participating in the task based on a central AUV, while the low-level component reduces exposure probabilities through power and trajectory control by the participating AUVs. Simulation results show that the proposed framework achieves rapid convergence, outperforms benchmark algorithms in terms of performance, and maximizes long-term cooperative efficiency while ensuring covert operations.

\end{abstract}


\IEEEpeerreviewmaketitle

\section{Introduction}
\IEEEPARstart{U}{nderwater} operations encompass tasks such as military reconnaissance, maritime security, marine surveying, and ecological monitoring, all of which are crucial for national security and the blue economy. Among these, underwater acoustic communication and collaborative sensing are key capabilities that support agents (AUVs/UUVs/sensors) in completing their missions \cite{acoustic}. However, in military or security-sensitive scenarios, the demand for covert operations is equally stringent: once communication or maneuvers are detected by adversaries, the platform and mission face the risk of exposure and countermeasures. Existing research indicates that underwater acoustic spreading, power control, and trajectory avoidance can achieve low intercept/low detection probability (LPI/LPD) under low signal-to-noise ratios \cite{covert01} \cite{covert02}, but their covert boundaries are limited by channel conditions and environmental sound fields. At the same time, an increase in the number of participating agents significantly enhances exposure risks, thus giving rise to the ``who to deploy, when to deploy, and how to deploy" collaboration-covert dilemma.

Existing work largely focuses on covert communication technologies for single agents \cite{power}, or neglects the covert constraints in collaborative path planning or resource allocation for multi-agent tasks such as coverage, searching \cite{search}, and tracking \cite{formation} \cite{track}, making it difficult to meet practical application needs. Methodologically, traditional distributed control, optimization, and game-theoretical approaches can achieve collaboration to some extent \cite{distribute}, but they rely on precise models and prior information, making them challenging to adapt to complex and dynamic underwater environments. Heuristic and evolutionary algorithms \cite{Heuristic}, while flexible, often struggle to meet real-time requirements. Deep reinforcement learning, due to its adaptability in various environments, has been applied to address several underwater issues. 

However, flat (single-layer) structures are primarily designed for short-term, local decision-making and struggle to handle high-dimensional, long-horizon, and multi-constraint problems. In contrast, hierarchical reinforcement learning is more suitable for decoupling global tasks from local control, balancing long-term benefits. Based on this, this paper proposes a dual time-scale hierarchical multi-agent proximal policy optimization (H-MAPPO) framework that maximizes the collaborative efficiency while ensuring compliance with covert and energy constraints.

\theoremstyle{Observation}
\newtheorem{observation}{\textit{Observation}}
\theoremstyle{Lemma} 
\newtheorem{lemma}{{\textit{Lemma}}}

\section{System Model and Problem Formulation}
\subsection{Scenario Description}
\begin{figure}[h]
	\centering
	\captionsetup{font={small}}    \includegraphics[width=0.86\columnwidth]{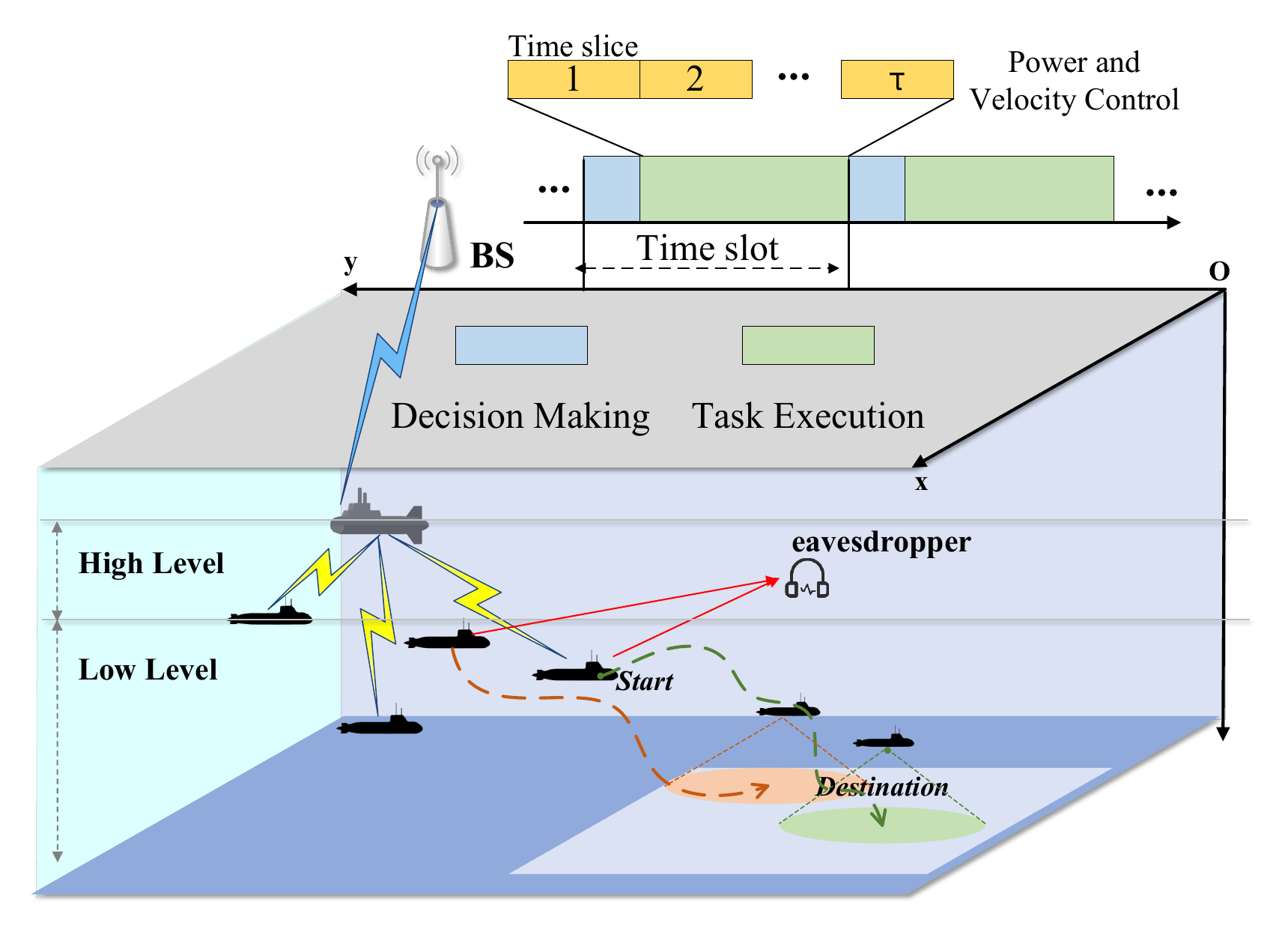}
	\caption{Network scenario and time slot structure, where the red arrow indicates the eavesdropping links.}
	\label{system}
\end{figure}
We construct a multi-AUV network scenario for underwater cooperative exploration tasks, as illustrated in Fig.~\ref{system}. The participants include a central control AUV and a cooperative team consisting of $N$ AUVs. Meanwhile, the network contains a potential eavesdropper located at a fixed position. The shore-based station issues a command to the central AUV, specifying the target exploration area as $([x,y,z], l, w)$. Upon receiving the command, the central AUV performs task planning, i.e., selecting appropriate AUVs from the team to participate in the current task and assigning instructions to them. The selected AUVs then autonomously plan their paths to the target area and carry out the exploration task.

The entire task execution process exhibits an explicit hierarchical decision-making structure. First, after each task command is issued, the central AUV performs a task assignment. Subsequently, the designated AUVs conduct continuous action control throughout the execution process. Therefore, we structure the whole task with high-level decision-making over time slots $t$ and low-level execution over time slice $\tau$. One time slot represents a complete cycle of accomplishing a task. In each $t$, the central AUV needs to make a decision, which is represented by the variable $G(t) \!=\! \{G_1(t), \ldots, G_N(t)\}$, where $G(t)$ is a binary indicator vector. Specifically, for a given $n, \ \forall n \in [1, N]$, $G_n(t)\!=\!1$ indicates that $n$th AUV participates in the task, otherwise $G_n(t)\!=\!0$ holds.
Furthermore, each task is decomposed into consecutive time slices $\tau$. That is, after the central AUV selects a group of AUVs to participate in the task, the execution process of this group is further divided into $T$ sequential decision-making stages. Each AUV achieves covert and efficient task completion by controlling its transmit power $P_n(t,\tau)$ and three-dimensional velocity vector $V_n(t,\tau)$. To ensure the effectiveness of the exploration task, each AUV hovers in the designated area for real-time monitoring once it arrives, until the current task cycle is completed.

The positions of the $N$ cooperative AUVs are denoted as ${p}_n(t,\tau)$, the central AUV's position as ${p}_c(t,\tau)$, and the eavesdropper’s location is fixed as ${p}_e$. Accordingly, the Euclidean distance between the central AUV and the $n$-th AUV is $d_{c,n}(t,\tau) = \|{p}_n(t,\tau) - {p}_c(t,\tau)\|_2$,
while the distance between any two AUVs is
$d_{i,j}(t,\tau) = \|{p}_i(t,\tau) - {p}_j(t,\tau)\|_2$, $\forall i,j \in N$.
Similarly, the distance from the $n$-th AUV to the eavesdropper is expressed as $d_{n,e}(t,\tau) = \|{p}_n(t,\tau) - {p}_e\|_2$.
The velocity of the $n$th AUV is denoted by $\vec{v}_n$. Hence, its position is updated over time slices according to ${p}_n(t,\tau) = {p}_n(t,\tau-1) + \tau \vec{v}_n$.

\subsection{Underwater Acoustic Communication Model}
Underwater acoustic waves are the key technology for achieving long-distance wireless communications. However, the signals experience severe attenuation during propagation, which directly affects the effectiveness of cooperation. 
According to the Thorp model \cite{Thorp}, underwater acoustic signals suffer from both spreading loss and frequency-dependent absorption loss during transmission, and the overall path loss can be expressed as $A(f,d) = d^k 10^{\frac{\alpha(f)  d}{10}}$, where $d$ is the propagation distance and $k$ is the diffusion factor. Based on this, the channel gain modeling is obtained as 
\begin{equation}
    g_{i,j}(t,\tau) = 1/A(f,d_{i,j}(t,\tau)).
\end{equation}

In addition to signal attenuation, environmental noise also affects the quality of communication. Underwater acoustic channel noise exhibits frequency correlation and is the result of the superposition of multiple sources, including turbulence, oceanic conditions, ships, and thermal noise. The specific noise calculation formula is given by
\begin{equation}\label{noise}
\begin{aligned}
&10\lg \mathcal{N}_{t}(f) \!=\! 17 - 30\lg f, \\
&10\lg \mathcal{N}_{s}(f) \!=\! 30 + 20s + 26\lg f - 60\lg(f + 0.03), \\
&10\lg \mathcal{N}_{w}(f) \!=\! 50 \!+\! 7.5w^{1/2} \!+\! 20\lg f \!-\! 40\lg(f \!+\! 0.4), \\
&10\lg \mathcal{N}_{th}(f) \!=\! -15 + 20\lg f.
\end{aligned}
\end{equation}

Therefore, the total noise is calculated as $\mathcal{N}(f) = \mathcal{N}_t(f)+\mathcal{N}_s(f)+\mathcal{N}_w(f)+\mathcal{N}_{th}(f)$. Here, $s$ represents the navigation activity factor, and $w$ indicates the wind speed.
\subsection{Signal Model}
We assume that the AUV employs orthogonal multiple access technology, allowing multiple AUVs to transmit data to the central AUV simultaneously without interfering with each other. In the time slice $\tau$, the signal received by the AUV from the central AUV is denoted as
\begin{equation}
    {{y}_{n}}\left( t,\tau  \right)\!=\!
    \left\{ 
\begin{aligned}
  & {\sqrt{{{P}_{n}}\left( t,\tau  \right){{g}_{n,AUV}}\left( t,\tau  \right)}}{{s}_{n}}\!+\!{{N}_{n}}(t),\;\;\;\;{\mathcal{K}_{1}},\\ 
 & {{N}_{n}}(t),\;\;\;\;\;\;\;\;\;\;\;\;\;\;\;\;\;\;\;\;\;\;\;\;\;\;\;\;\;\;\;\;\;\;\;\;\;\;\;\;\;\;\;\;\;\;\;{\mathcal{K}_{0}}, \\ 
\end{aligned} \right.
\end{equation}
where $s_n$ and $P_n$ represent the transmission power and signal of the AUV, respectively. $\mathcal{K}_1$ indicates the presence of communication, while $\mathcal{K}_0$ indicates the absence of communication. The received signal at the eavesdropper is calculated as
\begin{equation}
    {{y}_{e}}\left( t,\tau  \right)\!=\!\left\{ 
    \begin{aligned}
  & \sum\limits_{n\in N}{\sqrt{{{P}_{n}}\left( t,\tau  \right){{g}_{n,e}}\left( t,\tau  \right)}}{{s}_{n}}\!+\!{{N}_{e}}(t),\;\;\;{\mathcal{K}_{1}}, \\ 
 & {{N}_{e}}(t),\;\;\;\;\;\;\;\;\;\;\;\;\;\;\;\;\;\;\;\;\;\;\;\;\;\;\;\;\;\;\;\;\;\;\;\;\;\;\;\;\;\;\;\;\;\;\;\;{\mathcal{K}_{0}}. \\ 
\end{aligned} \right.
\end{equation}
here, $\mathcal{N}_n$ and $\mathcal{N}_e$ represent the noise, which follows a Gaussian distribution. Additionally, the signal $s_n$ from different AUVs is incoherent at the eavesdropper, so the received signal at the eavesdropper can be considered as the superposition of each signal component along with the noise. Therefore, the signal-to-noise ratio (SNR) can be modeled as
\begin{equation}
    {{\gamma }_{e}}\left( t,\tau  \right)\!=\!\sum\limits_{n\in \mathcal{N}}{\frac{G_{n}^{2}\left( t \right){{P}_{n}}\left( t,\tau  \right)}{{{A}_{n,e}}\left( t,\tau  \right){\mathcal{N}_{e}}(t)}},
\end{equation}
The data rate between AUVs $m$ and $n$ is given by
\begin{equation}
    {{R}_{m,n}}\left( t,\tau  \right)\!=\!B{{\log }_{2}}\left( 1\!+\!\frac{{{\sum\limits_{m,n\in N}{G_{m}^{2}\left( t \right){{P}_{m}}\left( t,\tau  \right){{g}_{m,n}}\left( t,\tau  \right)}}}}{\mathcal{N}_n\left( t \right)} \right).
\end{equation}

\subsection{Covertness Constraint}
To ensure that the team of AUVs remains undetected by potential eavesdroppers during collaborative task execution, their communication behavior must satisfy strict covertness constraints. In this paper, we employ the Kullback–Leibler (KL) divergence from information theory to quantitatively assess the system’s covertness. Specifically, KL divergence measures the difference between the probability distributions $\mathcal{P}_0$ and $\mathcal{P}_1$ of the signals received by the eavesdropper under two hypotheses ($\mathcal{K}_0$ and $\mathcal{K}_1$). A lower KL divergence indicates that the two distributions are more difficult to distinguish, implying greater covertness. 

Therefore, the system can satisfy covertness requirements by controlling the transmission power of the AUVs. By introducing the KL divergence $D(\mathcal{P}_0 \| \mathcal{P}_1)$, a mathematically tractable covertness constraint can be constructed. The KL divergence can be precisely expressed as a function of the eavesdropper’s signal-to-noise ratio (SNR) \cite{power}, i.e.,
\begin{equation}
    D\left( {\mathcal{P}_{0}}||{\mathcal{P}_{1}} \right)=\frac{1}{2}\left(\ln \left( 1+{{\gamma }_{e}}\left( t,\tau  \right) \right)-\frac{{{\gamma }_{e}}\left( t,\tau  \right)}{1+{{\gamma }_{e}}\left( t,\tau  \right)} \right).
\end{equation}

\subsection{Task Model}
We divide the entire task process into four phases: the command dispatch phase, the AUV mobility phase, the task execution phase, and the data transmission phase. \hspace{-2mm}
\begin{itemize}
    \item In the command dispatch phase, the time delay includes data transmission delay and underwater transmission latency, which is calculated as 
\begin{equation}
    T_n^d(t)=\frac{v_u}{d_{n,AUV}(t)}+\frac{d(t)}{R_{n,AUV}(t)},
\end{equation}
where $d(t)$ is the amount of data for the commands sent by the central AUV at time slot $t$, and $v_u$ represents the transmission speed of underwater acoustic waves.
\item In the AUV mobility phase, each AUV plans its optimal path to reach the target area based on its current status, such as energy, position, and so on. Considering that the target area is a rectangular region, the exact destination coordinates are unknown for each AUV. However, finding the precise target location coordinates is essential for effective navigation. Assuming the exploration area of the AUV is circular, we can simplify the problem by covering the rectangular area with as many non-overlapping circles as possible.

Here, we employ a greedy algorithm, which randomly assigns positions to each AUV in a cycle. The process continues as long as the coverage area of the current AUV does not significantly overlap with that of the previous AUV, until all AUV positions are assigned. Since each AUV has different computational capabilities, given that the area of its exploration zone is related to its computational ability, which we model as $r_n=r_b+\delta \ln \left(\frac{C_n}{C}+1\right)$, where $r_b$ is the basic exploration radius, $C$ is the threshold for reference computational ability. 

Therefore, the task coverage rate can be defined as
\begin{equation}
    \varsigma(t)=\frac{\sum_{n \in N} G_n(t)\pi r_n^2}{L(t)^2}.
\end{equation}

\item Next, we model the time delay throughout the AUV mobility process. Considering that underwater currents have a certain velocity, we refer to several numerical models of Lamb vortices and the Navier-Stokes equations \cite{Ocean_current01} \cite{Ocean_current02} to simulate ocean turbulence in three-dimensional space. We define this velocity as $\overset{\to }{\mathop{V}}_{T}$, which allows us to obtain the relative velocity between the AUV and the current:
${\overset{\to }{\mathop{V}}}_n^{'}\left( t,\tau  \right)={{\overset{\to }{\mathop{V}}_{n}}}\left( t,\tau  \right)-{{\overset{\to }{\mathop{V}}_{T}}}.$
The time delay for the AUV to move from its current position to the assigned task coordinate can be modeled as
\begin{equation}
\begin{split}
        &T_{n}^{m}(t) =\Delta \tau \\& \min \left( \inf \{ t \in \mathbb{N}^+ \mid \| {{\overset{\to }{\mathop{p}}\,}_{n}}\left( t,\tau  \right)\!-\!{{\overset{\to }{\mathop{p}}\,}_{sub,n}}\left( t \right) \| \!\le\! r_n \} \right).
    \label{move}
\end{split}
\end{equation}
After the AUV arrives at the target area, it scans the specified location and completes data collection. The time delay for this process is given by
$T_{n}^{e}=\frac{2\pi }{\varpi }$,
where $\varpi$ is the effective width covered by the AUV's sonar during a single rotation scan.

\item In the task execution phase, the data collected by the AUV is: $D_{n}^{'}\left( t \right)=\varphi \left( t \right) \pi {{r}_{n}}^{2},$ $\varphi$ represents the amount of data collected in a single collection process. The time delay required to upload to the AUV is modeled similarly to the command dispatch phase, i.e.,
\begin{equation}
    T_{n}^{up}\left( t \right)=\frac{D_{n}^{'}\left( t \right)}{{{R}_{n,AUV}}\left( t,\tau  \right)}+\frac{v_u}{d_{n,AUV}(t,\tau)}.
\end{equation}
Finally, the total task execution delay is determined by the delay of the slowest AUV executing the task:
\begin{equation}
    {{T}_{task}}\left( t \right)\!=\!\underset{n\in N}{\mathop{\max }}\,\left( T_{m}^{d}\left( t \right)\!+\!T_{n}^{m}\left( t \right)\!+\!T_n^e\!+\!T_{n}^{up}\left( t \right) \right).
\end{equation}
\end{itemize}

\subsection{Energy Consumption Model}
The energy consumption of the AUV is divided into three components: mobility energy, exploration energy, and data upload energy. 

\begin{figure*}[ht]
    \centering
    \begin{equation}
        \begin{aligned}     \label{E_m_sum}
    & E_{n}^{m}\left( t,\tau  \right)=E_{n}^{h}\left( t,\tau  \right)+E_{n}^{d}\left( t,\tau  \right)+E_{n}^{f}\left( t,\tau  \right) \\ 
     & \text{         =}\frac{\frac{{{G}^{2}}\Delta\tau }{\sqrt{2}A{{\rho }_{L}}}}{\sqrt{\left( p_{n}^{x}{{\left( t,\tau  \right)}^{2}}+p_{n}^{y}{{\left( t,\tau  \right)}^{2}} \right)+{{\left( p_{n}^{x}{{\left( t,\tau  \right)}^{2}}+p_{n}^{y}{{\left( t,\tau  \right)}^{2}} \right)}^{2}}+\frac{{{G}^{2}}}{{{A}^{2}}\rho _{L}^{2}}}}+Gp_{n}^{z}\left( t,\tau  \right)\Delta\tau +\frac{1}{2}{A{{C}_{d}}{\rho }_{L}}\Delta\tau \left\| \overset{\to }{\mathop{V}}\,_{n}^{'}\left( t,\tau  \right) \right\|_{2}^{3} \\ 
    \end{aligned}
    \end{equation}
    \noindent\rule{1\linewidth}{0.4pt}
\end{figure*}

\begin{itemize}
    \item  \textbf{Mobility energy consumption}. It is based on the current model and refers to the energy expended when moving within the current. In the modeling process, it is decomposed into horizontal, vertical, and fluid drag energy consumption, which is modeled as (\ref{E_m_sum}). The key physical parameters are explained as follows. $G$ represents the gravitational force due to the AUV's mass, $\rho$ is the fluid density of the seawater medium, and $A$ is the AUV's cross-sectional area during vertical movement, which affects its vertical propulsion efficiency. $C_d$ is the dimensionless drag coefficient, which quantifies the magnitude of drag resistance the AUV experiences while moving through the water.
    \item \textbf{Exploration energy consumption}. It depends on the task and the sensor performance: $E_n^{d}=\pi r_n^2 \kappa$. 
    \item \textbf{Upload energy consumption}. It is the energy used by the AUV to transmit the data to the central AUV through the acoustic channel, which is determined by the transmission power and transmission delay: $E_n^{u}(t)=\frac{P_n(t,\tau)}{\Upsilon}  \cdot \frac{D_n(t)}{R_{n,AUV}(t,\tau)}$, $\Upsilon$ is the conversion efficiency. Let the initial energy of the AUV be $E_n^{i}$, so the remaining energy is given by $E_n(t)\!=\!E_n^{i}\!-\!E_n^m(t)\!-\!E_n^d(t)\!-\!E_n^u(t)$.
\end{itemize}

\subsection{Problem Formulation}
To assess the efficiency of completing the collective task, we define the AUV collaboration efficiency, which is related to the task coverage rate and the total delay:
\begin{equation}
    \eta \left( t \right)=\frac{\varsigma \left( t \right)}{{{T}_{task}}\left( t \right)}.
\end{equation}
Therefore, let $\boldsymbol{\Theta}\!=\!\{P\left( t,\tau  \right),V\left( t,\tau  \right),G\left( t \right)\}$, an optimization problem is formulated to maximizing collaboration efficiency:
\begin{align}
  & \underset{\boldsymbol{\Theta}}{\mathop{\max }}\,\text{ }\eta , \label{origin_problem}\\ 
 & s.t.\;\;\;\text{    }D\left( {\mathcal{P}_{0}}||{\mathcal{P}_{1}} \right)\left( t,\tau  \right)\le 2{{\epsilon_c }^{2}},\tag{\ref{origin_problem}a} \\ 
 & \;\;\;\;\;\;\;\;\text{        }{{P}_{\min }}\le {{P}_{n}}\left( t,\tau  \right)\le {{P}_{\max }},\forall n\in N, \tag{\ref{origin_problem}b}\\ 
 & \;\;\;\;\;\;\;\;\text{        }{{G}_{n}}\left( t \right)\in \left\{ 0,1 \right\},\forall n\in N,\tag{\ref{origin_problem}c} \\  
 & \;\;\;\;\;\;\;\;\text{        }{{E}_{n}}\left( t \right)\ge 0,\text{        }\forall n\in N. \tag{\ref{origin_problem}d}
\end{align}
The constraints include the communication covertness defined by (\ref{origin_problem}a), power limitations (\ref{origin_problem}b), the binary AUV selection decision (\ref{origin_problem}c), as well as (\ref{origin_problem}d) the energy consumption of the AUVs as specified. Since the problem involves decision-making on two different time scales, with AUV selection needed during each time slot and power-speed control during each time slice, we propose a dual-timescale multi-agent reinforcement learning framework.

\section{The Proposed MADRL-Based Design}
\subsection{MDP Formulation}
To address the aforementioned optimization problem, this section provides a detailed description of our proposed dual-scale reinforcement learning framework. Specifically, the decision-making of the central AUV at each time slot is modeled as an MDP, while the movement strategy of each AUV at each time slice is modeled as a partially observable Markov decision process (POMDP).

For the MDP modeling of the central AUV decision-making phase, the essential elements include:
\begin{itemize}
    \item \textit{State}: The state observed by the central AUV is global, which includes the current positions of all AUVs and their remaining energy information, denoted as: $\mathcal{S}_{global}=\{p(t),E(t)\}$.
    \item \textit{Action}: The central AUV selects the AUVs that will participate in the current time slot task: $\mathcal{A}_{c}=\{G_n(t)\}$.
    \item \textit{Reward}: The reward function is used to comprehensively evaluate the performance over the entire task cycle, aiming to maximize cooperative efficiency. It is defined as: $\mathcal{R}_{global}=\Delta_1  \varsigma+\Delta_2  T_{task}+\Delta_3  avg(\mathcal{R})$.
\end{itemize}

After the central AUV determines the AUV group participating in the task, the decision-making process of each activated AUV $n$ is modeled as a POMDP.
\begin{itemize}
    \item \textit{State}: The complete physical information of the AUV includes: the current position in the time slot, operational speed, and remaining energy: $\mathcal{S}_n=\{p_n(t,\tau),V_n(t,\tau),E_n(t,\tau)\}$.
    \item \textit{Action}: The action selected by the AUV at time $(t, \tau)$ includes a transmission power and a three-dimensional thrust velocity vector: $\mathcal{A}_n=\{P_n(t,\tau),V_n(t,\tau)\}$.
    \item \textit{Observation}: Due to the limitations of perception and communication, the AUV can only obtain partial information. Besides its own physical information, it also includes the current ocean current velocity, the distance to the sub-target, and the distance to the central AUV: $\mathcal{O}_n=\{d_{n,c}(t,\tau),d_{n,sub}(t,\tau),p_n(t,\tau),V_n(t,\tau),E_n(t,\tau)\}$
    \item \textit{Reward}: To better guide the AUV's behavior at each control step, it is designed as a multi-objective weighted sum:
    $\mathcal{R}_{n}(t,\tau) = w_c  \mathbb{I}(D(\mathcal{P}_0||\mathcal{P}_1)(t,\tau) \le 2{\epsilon_c}^2) + w_p \Delta d_{n,sub}(t,\tau) + w_b \mathbb{I}(d_{n,sub}(t,\tau) < r_n) - w_e \text{ReLU}(-E_{n}(t,\tau)).$
    The first part is the covertness reward, which provides rewards or penalties based on whether the total $KL$ divergence of the team meets certain constraints. Additionally, a reward is given for the reduction in the distance $\Delta d$ between the AUV and the target point, providing a dense directional cue. A one-time high reward is granted when the AUV first reaches the target area. Finally, to conserve energy, a penalty is imposed on AUVs that exceed their energy limits.
\end{itemize}

\subsection{PPO Solution}
We use the proximal policy optimization (PPO) algorithm to train the policy network $\pi_{\theta_{c}}$ of the central AUV. PPO updates the parameters $\theta_c$ of the actor network by optimizing a clipped surrogate objective function, which ensures sampling efficiency while limiting the step size of each update to prevent policy collapse. The core objective function is as follows:
\begin{equation}
    L^{\text{CLIP}}(\theta_c) \!=\! 
    \mathbb{E}_k \left[ 
    \min \left( r_t(\theta_c) A_t,
    \operatorname{clip}\big(r_t(\theta_c),1\!-\!\epsilon,1\!+\!\epsilon\big) A_t
    \right) 
    \right]
\end{equation}

The term $r_t({\theta}_c)$ represents the ratio between the new and old policies, $A_t$ is the advantage function, and $\epsilon$ is a hyperparameter used for clipping. The advantage function $A_t = G_t - V(\mathcal{S}_{global}(t))$, where $G_t$ is the cumulative return. $A_t$ is used to evaluate the effectiveness of the selected action relative to the average level, and it is specifically calculated using Generalized Advantage Estimation (GAE). Additionally, we employ an independent value network (critic) to assist in the computation of GAE, training it by minimizing the mean squared error between the predicted value and the actual cumulative reward, given by $(G_t - V_{{\phi}_c}(\mathcal{S}_{global}(t))^2$. Furthermore, both the actor and critic networks of the central AUV are updated using the Adam optimizer.

\begin{algorithm}[thb]
\SetAlgoLined
\small
\KwResult{Trained policies: central AUV's actor $\pi_{c}$ and AUVs' actors $\{\pi_i\}_{i=1}^N$.}
Initialize central AUV's networks $\pi_{c}(\theta_{c})$, $V_{c}(\phi_{c})$\;
Initialize AUVs' networks $\{\pi_i(\theta_i), V_i(\phi_i)\}_{i=1}^N$ with centralized critics\;
Initialize experience buffers $\mathcal{B}_{c}$ and $\mathcal{B}_{AUV}$\;

\For{episode = 1 \KwTo max\_episodes}{
    Reset env and get initial high-level state $\mathcal{S}_1$\;
    \For{high level timestep t = 1 \KwTo high\_level\_steps}{
        Select AUV schedule $G(t) \sim \pi_{AUV}(\mathcal{S}_{global}(t))$\;
        Initialize low-level env, get initial local observations $\mathbf{o}_1 = \{o_1^i\}_{i=1}^N$\;
        
        \For{$\tau$ = 1 \KwTo low\_level\_steps}{
            For each selected AUV $i \in G(t)$, select action $a_\tau^i \sim \pi_i(o_\tau^i)$\;
            Execute joint action $\mathcal{A}_i(\tau) = \{a_\tau^i\}$, observe rewards $\mathcal{R}_i(\tau)$ and next obs $\mathcal{O}_{global}({\tau+1})$\;
            Store transition $(\mathcal{O}_{global}(\tau+1), \mathcal{A}_i(\tau), \mathcal{R}_i(\tau))$ into $\mathcal{B}_{AUV}$\;
        }
        
        Observe next high-level state $\mathcal{S}(t+1)$ and high-level reward $\mathcal{R}(t)$\;
        Store $(\mathcal{S}(t), {G}(t), \mathcal{R}(t), \mathcal{S}(t+1))$ into $\mathcal{B}_{c}$\;
        
        \If{time to update AUVs}{
            Update all AUV networks $\{\pi_i(\theta_i), V_i(\phi_i)\}_{i=1}^N$ using PPO on data from $\mathcal{B}_{AUV}$\;
            Clear $\mathcal{B}_{AUV}$\;
        }
        
        \If{time to update central AUV}{
            Update central AUV networks $\pi_{c}(\theta_{c}), V_{c}(\phi_{c})$ using PPO on data from $\mathcal{B}_{c}$\;
            Clear $\mathcal{B}_{c}$\;
        }
    }
}
\caption{The Proposed H-MAPPO Algorithm}
\label{alg:hmapoo_compact}
\end{algorithm}

To address the low-level cooperative control issues of AUVs, we employ the MAPPO algorithm, which is a direct extension of PPO in the multi-agent domain. The update method for each AUV's actor network $\pi_{\theta_n}$ is the same as that of standard PPO, which maximizes the $L^{CLIP}(\theta_n)$ objective function. The key difference lies in the calculation of the advantage function $A_n$, which is based on a shared, centralized Critic whose input is the global state $V_{\Phi}(\mathcal{O}_{global})$. $\mathcal{O}_{global}$ is the concatenation of the observation vectors of all AUVs.  This allows for the evaluation of the current team status from a global perspective. The parameters $\phi$ are trained by minimizing the value prediction error at the joint state, given by $ (G_{t, \tau} -V_{\phi}(\mathcal{O}_{global}(t, \tau)))^2$. When calculating the advantage function for each AUV $A_n(t, \tau)$, the value estimates used are all derived from this centralized Critic.

Additionally, the proposed H-MAPPO adheres to the paradigm of centralized training and decentralized execution. During the training phase, a centralized critic network is used to gather and utilize information. In the execution phase, the actor networks of the AUVs make independent decisions based solely on the information they each obtain. This architecture significantly enhances learning efficiency and the performance of the final policy, as illustrated in \textbf{Algorithm \ref{alg:hmapoo_compact}}.

\section{Numerical Results and Discussion}
In this section, we evaluate the performance of the proposed framework through simulation experiments. The simulation scenario is a three-dimensional underwater environment measuring $200$m by $200$m and with a depth of $200$m. It includes a central AUV located at (0,0,-20), several moving AUVs, and a stationary eavesdropper located at (70,70,-10). The moving AUVs start randomly within the domain.  The specific simulation parameters are shown in Table \ref{parameters1} and \ref{hyperparams}.  
\begin{table}[thb]
	\centering
    \captionsetup{font={small}}
	\caption{System model Parameters}
	\label{parameters1}
	\small
	\begin{tabular}{l | l || l | l}
		\noalign{\global\arrayrulewidth=0.3mm}
		\hline
		\textbf{Parameter}  &\textbf{Value } &\textbf{Parameter}  & \textbf{Value } \\
		\hline
		 $N$   & 5 &  $E^{i}_n$   &$10000\sim 20000$\\
		 $P_{min}$  &0.01 &   $P_{max}$&  $2$\\
         $r_b$  &5  &    $C$&   10 \\
		$A$  & 0.1 & $C_d$ & 0.8 \\
        $\Delta \tau$     &2s & $L$        &0.5\\   
		 $C_m$ &5& $B$    &     10 MHz\\
		$k$ &1.5 & $V_{max}$ &5 m/s\\       
		$F_{u,c}$   &50 GHz    &  $f$ &    30 kHz\\
        $\epsilon_c$   &0.05    &  $r_b$ &   5 m\\
		\noalign{\global\arrayrulewidth=0.3mm}
		\hline
	\end{tabular}
\end{table}	

\begin{table}[thb]
\centering
\captionsetup{font={small}}
\caption{Network Parameters}
\label{hyperparams}
\small
\begin{tabular}{l | c || l | c}
\noalign{\global\arrayrulewidth=0.3mm}
\hline
\textbf{Parameter} & \textbf{Value} & \textbf{Parameter} & \textbf{Value} \\
\hline
Episodes & 2000 & High Level Steps & 10 \\
Low Level Steps & 100 & $LR_{\text{actor}}$ & $3 e{-5}$ \\
$LR_{\text{critic}}$ & $5 e{-5}$ & $\gamma$ & 0.99 \\
$\varepsilon$ & 0.2&  $\lambda_{\text{GAE}}$ & 0.95\\
Batch Size(AUV) & 512 & Batch Size(c) & 16 \\
\noalign{\global\arrayrulewidth=0.3mm}
\hline
\end{tabular}
\end{table}

\begin{figure}
    \centering
    \includegraphics[width=0.95\linewidth]{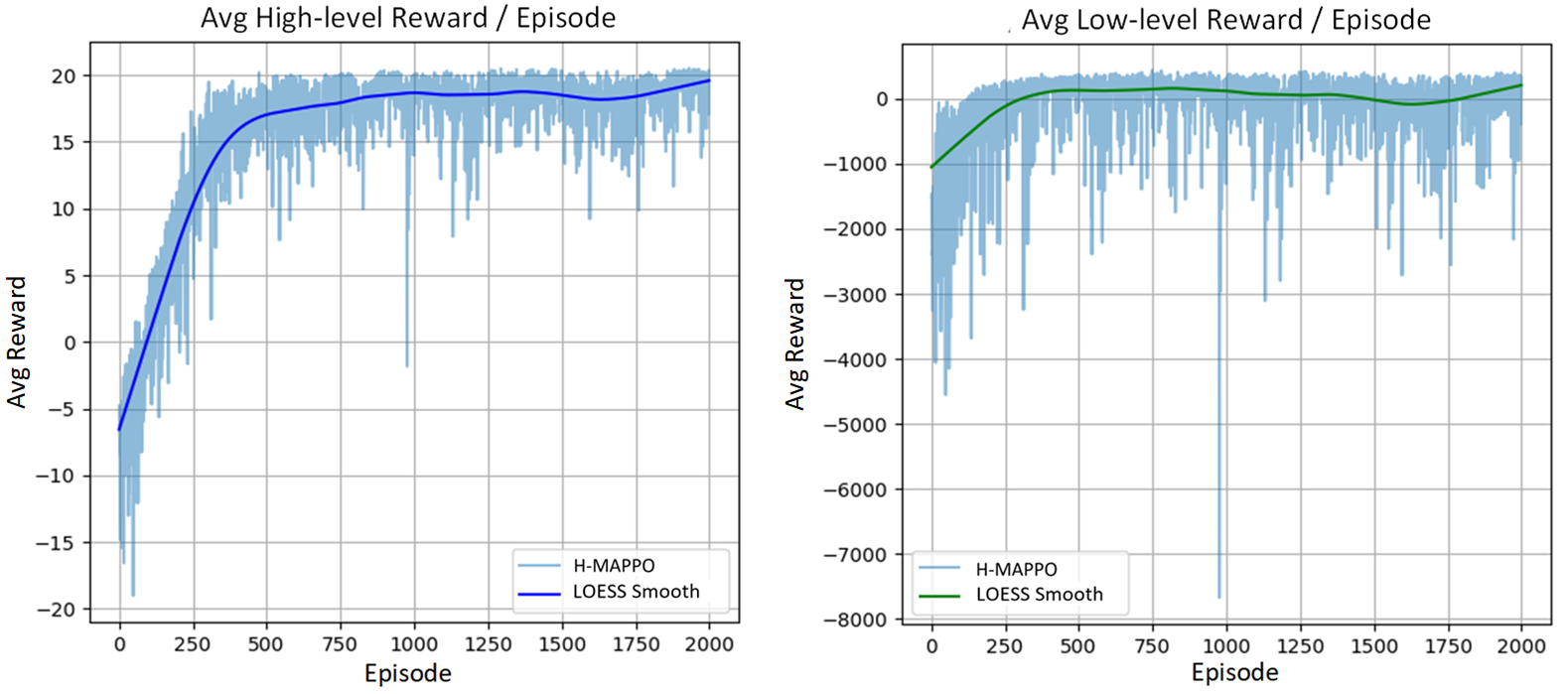}
    \captionsetup{font={small}}
    \caption{Convergence performance of high-level and low-level average reward.}
    \label{train}
\end{figure}

First, we verify the convergence of the H-MAPPO framework. As shown in Fig.~\ref{train}, both the average high-level reward and low-level reward can achieve stable convergence.

\begin{figure}
    \centering
    \includegraphics[width=1\linewidth]{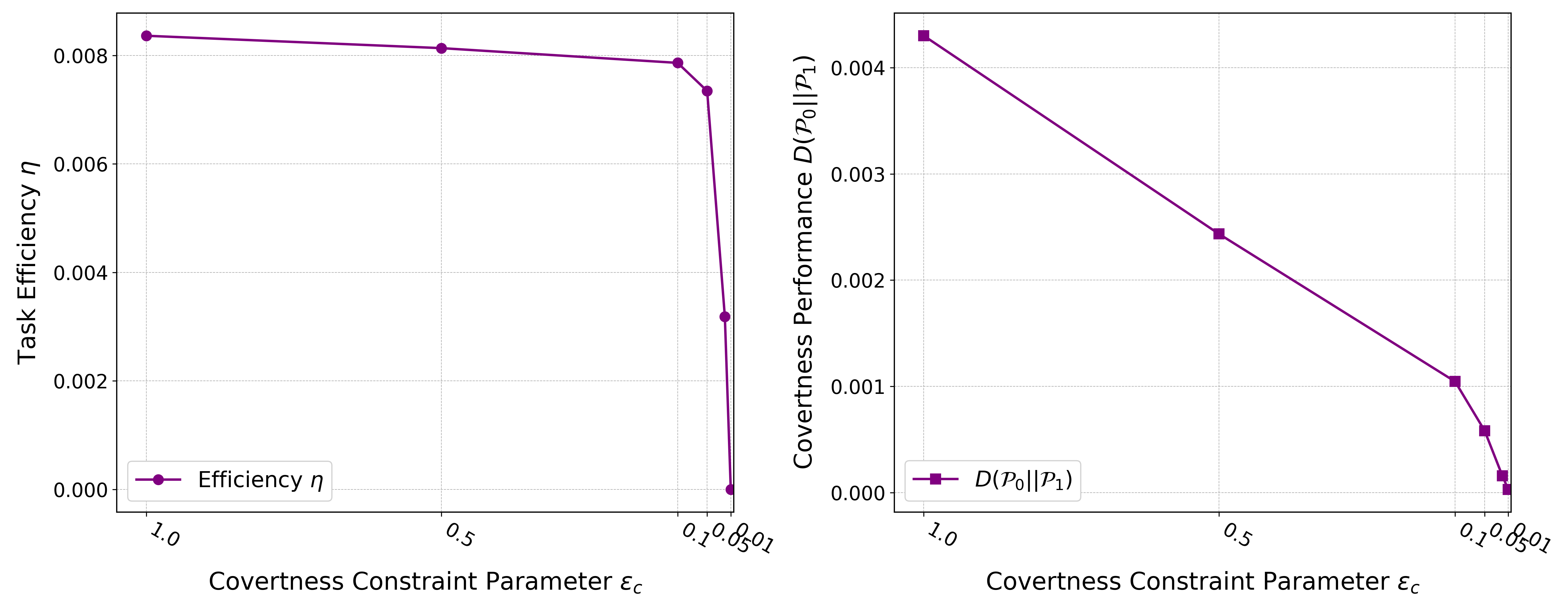}
    \captionsetup{font={small}}
    \caption{System performance with varying covertness constraint parameter.}
    \label{covertness}
\end{figure}

To explore the adaptability of our framework under different covert requirements, we adjusted the covertness constraint parameters and observed their impact on the optimization objectives of collaborative efficiency $\eta$ and covertness performance. The left figure in Fig.~\ref{covertness} demonstrates that as the covertness constraints tightened, the task efficiency correspondingly decreased. This is because stricter covertness requirements compel AUVs to employ lower transmission power and more stringent trajectory planning, and even to make certain sacrifices, such as reducing the number of participating AUVs or choosing detour paths farther away from potential eavesdroppers. The right side of Fig.~\ref{covertness} demonstrates that our algorithm effectively adheres to the imposed covertness constraints, with the covertness performance exhibiting a strictly monotonic decrease as the covertness parameter decreases.

\begin{figure}[t]
    \centering    \includegraphics[width=0.9\linewidth]{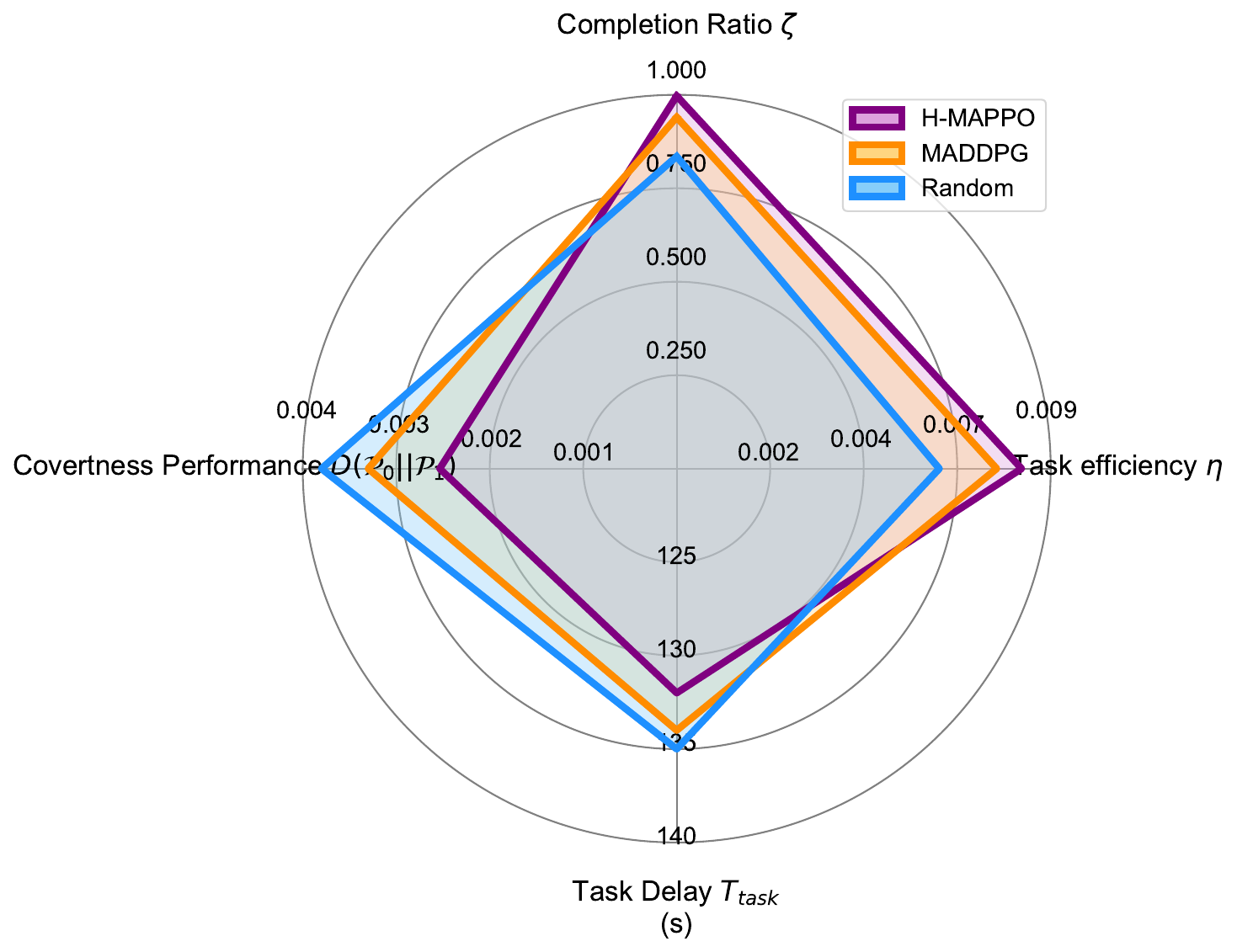}
    \captionsetup{font={small}}
    \caption{Overall performance evaluation.}
    \label{radar}
\end{figure}

Fig.~\ref{radar} provides an intuitive comparison of our proposed H-MAPPO framework against MADDPG and a random AUV delegation strategy across four key dimensions: cooperation efficiency, task completion ratio, covertness, and task efficiency. The results show that our method achieves the best performance in all dimensions. While MADDPG demonstrates some learning capability and outperforms the random strategy, it falls behind our approach in balancing task efficiency and covertness, as its flat structure remains inadequate for handling long-horizon, multi-constraint problems. The random strategy performs the worst across all dimensions, particularly in covertness control, likely because its random selection of participating AUVs fails to account for their relative positions to the eavesdropper, or because dispatching too many AUVs increases the overall exposure risk of the team.
\section{Conclusion}

This paper examines the trade-off between collaborative efficiency and covert communication in underwater multi-AUV cooperative tasks by introducing a novel dual-timescale H-MAPPO framework. 
We optimize policies using the PPO and MAPPO algorithms at both levels, guiding the AUVs to collaborate efficiently while ensuring they meet various constraints through carefully designed reward functions at each level. 
Simulation results demonstrate the effectiveness of the proposed framework, showing significant advantages over baseline algorithms in key performance indicators, such as collaborative efficiency and task completion rate, while successfully keeping the risks associated with covert communication below acceptable thresholds.


\ifCLASSOPTIONcaptionsoff
  \newpage
\fi

\vspace{12pt}


%

%
%
%




\end{spacing}
\end{document}